\pdfoutput=1
\documentclass[11pt]{article}

\usepackage{acl}

\usepackage{times}
\usepackage{latexsym}

\usepackage[T1]{fontenc}
\usepackage[utf8]{inputenc}

\usepackage{microtype}

\usepackage{amsmath}
\usepackage{pdfpages}
\usepackage{pgffor}
\usepackage{multirow}
\usepackage{booktabs}
\usepackage{color}
\usepackage{colortbl}
\usepackage{cleveref}
\crefformat{section}{\S#2#1#3} % see manual of cleveref, section 8.2.1

\definecolor{xing}{rgb}{1.0, 0.03, 0.0}

\title{MoK-RAG: Mixture of Knowledge Paths Enhanced \\Retrieval-Augmented Generation for Embodied AI Environments}

\author{
\bf{Zhengsheng Guo},~~
\bf{Linwei Zheng},~~
Xinyang Chen,~~
Xuefeng Bai,~~
Kehai Chen\thanks{~~Corresponding Author},~~
Min Zhang~~\\
\text Institute of Computing and Intelligence, Harbin Institute of Technology, Shenzhen, China \\
 \texttt{zhengshguo@gmail.com, 220110604@stu.hit.edu.cn} 
}

\begin{document}
\maketitle
\begin{abstract}
While human cognition inherently retrieves information from diverse and specialized knowledge sources during decision-making processes, current Retrieval-Augmented Generation (RAG) systems typically operate through single-source knowledge retrieval, leading to a cognitive-algorithmic discrepancy.
To bridge this gap, we introduce MoK-RAG, a novel multi-source RAG framework that implements a mixture of knowledge paths enhanced retrieval mechanism through functional partitioning of a large language model (LLM) corpus into distinct sections, enabling retrieval from multiple specialized knowledge paths.
Applied to the generation of 3D simulated environments, our proposed MoK-RAG3D enhances this paradigm by partitioning 3D assets into distinct sections and organizing them based on a hierarchical knowledge tree structure. 
Different from previous methods that only use manual evaluation, we pioneered the introduction of automated evaluation methods for 3D scenes. Both automatic and human evaluations in our experiments demonstrate that MoK-RAG3D can assist Embodied AI agents in generating diverse scenes.
\end{abstract}

\section{Introduction}
The rapid advancements in language models have led to the emergence of Retrieval-Augmented Generation (RAG) ~\citep{ji-etal-2024-rag}, which synergize the generative capacity of large language models (LLMs) ~\cite{gpt4oeval} with external knowledge retrieval. These systems substantially enhance the ability of LLMs to produce more accurate and contextually relevant responses by retrieving pertinent information from a predefined knowledge corpus. This approach has proven effective in various applications, including question answering and document generation, where grounding in external data improves output quality ~\cite{wang-etal-2024-rag}.
\begin{figure}[t]
	\centering
	\includegraphics[width=\columnwidth]{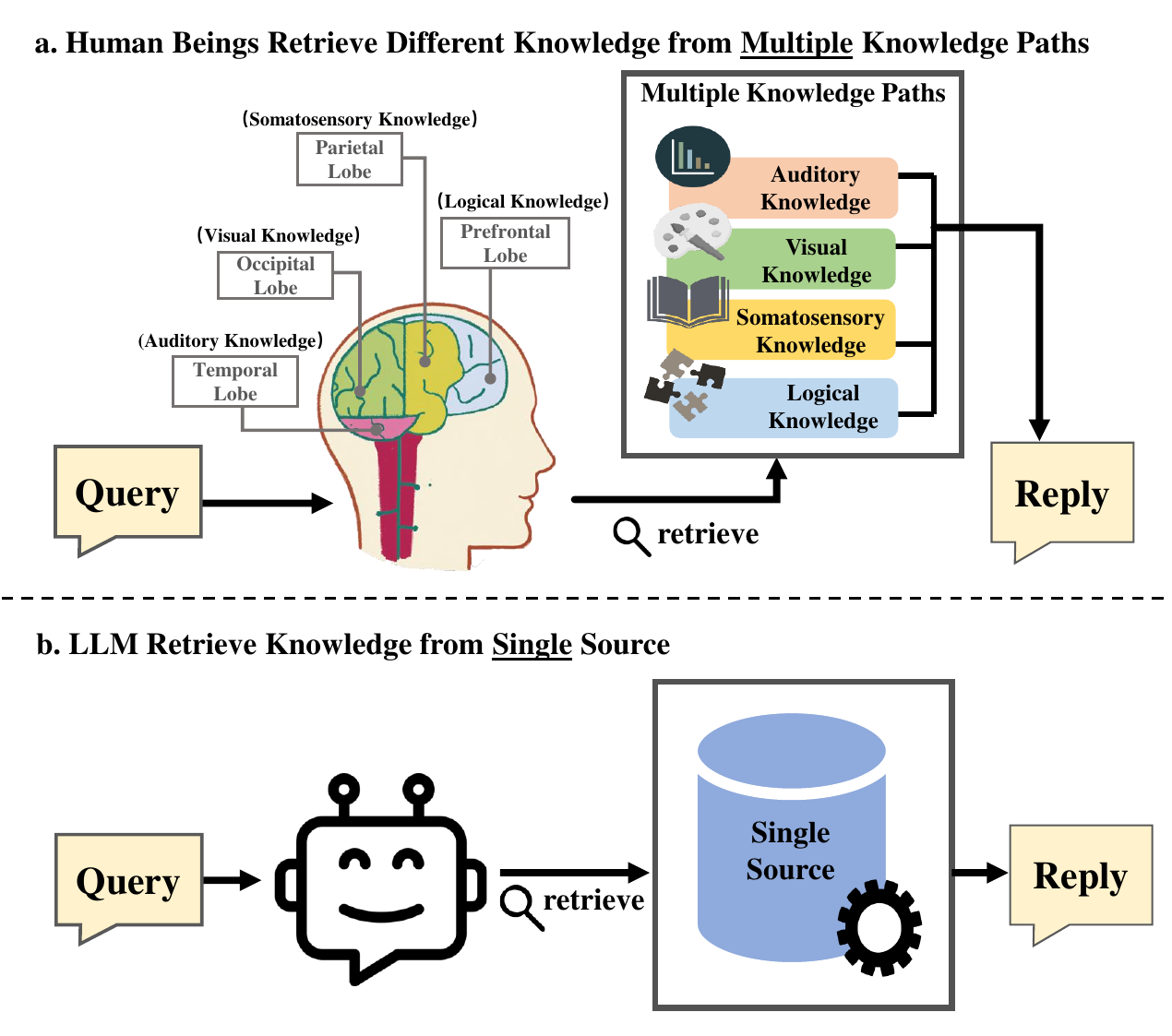}
	\caption{A figure showing the difference between human beings and LLM agents. In human cognition, decisions are often made by retrieving information from diverse knowledge sources. However, current Retrieval-Augmented Generation (RAG) systems typically rely on a single knowledge corpus.}
	\label{fig:motivation}
%	\vspace{-0.5em}
\end{figure}

While existing RAG systems demonstrate notable effectiveness, they remain fundamentally constrained by their reliance on a singular, generic knowledge corpus ~\cite{jiang2024longrag, sergent1987new}. 
This limitation prevents them from fully emulating the multifaceted and modular nature of human cognition, where decision-making inherently involves dynamic retrieval from multiple specialized knowledge sources. 
Neuroscientific studies by ~\citet{roland1998structural, gazzaniga1995principles, barrett2003human} reveal that human knowledge organization follows a specialized neural architecture: the left cerebral hemisphere predominantly processes analytical and logical information, while the right hemisphere specializes in creative synthesis and holistic pattern recognition. This neurocognitive division enables humans to perform context-sensitive information retrieval from distinct neural repositories when formulating responses to complex queries. 
Current RAG implementations, as illustrated in Figure~\ref{fig:motivation}, contrast sharply with this biological paradigm by operating through a monolithic knowledge base.
Such architectural simplicity inherently restricts their ability to perform domain-specific information retrieval and contextual adaptation, always leading to responses that are incomplete or lacking key details. We refer to this problem as \textit{Reply Missing}.

To address this problem, we propose MoK-RAG (Mixture of Knowledge Paths Enhanced Retrieval-Augmented Generation), a novel RAG framework that segments the LLM corpus into distinct sections, enabling simultaneous retrieval from multiple specialized knowledge paths.
Our approach models human cognitive specialization to enhance contextual relevance, adaptability, and mitigate the \textit{Reply Missing}.

Preliminary experiments indicate that the generation of 3D simulated environments, also exhibits a high occurrence rate of the \textit{Reply Missing} problem.
%To address these specialized requirements, we extend our framework through MoK-RAG3D, a novel approach that operationalizes MoK-RAG's core principles to enhance the generation of 3D environment.
Thus We extend MoK-RAG to propose MoK-RAG3D, a specialized adaptation designed to enhance the generation of 3D environments. 
MoK-RAG3D adheres to the MoK-RAG framework while introducing two domain-specific techniques:
%The methodology of MokRagFor3D obeys the MokRag framework and involves two primary steps. 
%First, Asset Partitioning is proposed as a systematic process for partitioning 3D assets into functionally discrete sections, based on their types and contextual relevance.
First, it splits 3D assets into distinct retrieval sections, categorizing them based on their types and contextual relevance. 
Second, Structural Knowledge Organization is utilized to organize these sections using a hierarchical knowledge tree structure, which facilitates efficient retrieval and assembly of assets, ensuring that the generated environments are both cohesive and contextually appropriate.
%Second, it organizes these sections using a hierarchical knowledge tree structure, which facilitates efficient retrieval and assembly of assets, ensuring that the generated environments are both cohesive and contextually appropriate. 
%To the best of our knowledge, MoK-Rag3D represents the first framework implementing multi-source retrieval mechanisms for the generation of 3D environments in embodied agent training. 
%Distinct from prior advanced approaches ~\citet{holodeck}, which only employ manual evaluation, we pioneered the introduction of automated evaluation methods for 3D scenes. 
%Both automatic and human evaluations in experiments demonstrate that MoK-RAG3D can facilitate Embodied AI agents in generating residential and diverse scenes. 
The contributions of our work can be summarized as follows: 
\begin{enumerate}
    \item  To address the \textit{Reply Missing} problem, we introduce MoK-RAG, the first multi-source RAG framework enabling multi-path knowledge retrieval.
    \item  To mitigate the high occurrence of \textit{Reply Missing} in 3D environment generation, we extend MoK-RAG to propose MoK-RAG3D.
    \item  MoK-RAG3D pioneers automated evaluation for 3D scene generation, with both automatic and human assessments confirming its effectiveness in enhancing Embodied AI agents' ability to generate diverse scenes.
\end{enumerate}

\section{Related Work}

\paragraph{Retrieval-Augmented Generation}
%RAG is one technical to enhances LLMs by retrieving relevant document chunks from external knowledge base through semantic similarity calculation. Previous Rag methods can be categorized into two groups: those focusing on retrieving algorithm~\citep{wang2024searching, jiang2024longrag, ji-etal-2024-rag, qian-etal-2024-grounding}, and those aiming to improve do better generation ~\citep{qi-etal-2024-model, wu-etal-2024-synchronous, fang-etal-2024-enhancing, adak-etal-2025-reversum, gou-etal-2023-diversify}. Moreover, RAG based LLM Agents is also popular~\citep{zhu-etal-2024-atm, wang-etal-2024-rag}. However, these methods put all items to one single corpus, not utilizing item's own feature to form multiple retrieving corpus. Based on that, we propose MokRag and to the best of our knowledge, MokRag is the first  multi-source  Rag Agent system.
Retrieval-Augmented Generation (RAG) enhances LLMs by retrieving relevant document chunks from external knowledge bases using semantic similarity. Existing RAG methods mainly focus on improving retrieval algorithms~\citep{wang2024searching, jiang2024longrag, ji-etal-2024-rag, qian-etal-2024-grounding} or refining generation quality~\citep{qi-etal-2024-model, wu-etal-2024-synchronous, fang-etal-2024-enhancing, adak-etal-2025-reversum, gou-etal-2023-diversify}. Additionally, RAG-based LLM agents have gained attention~\citep{zhu-etal-2024-atm, wang-etal-2024-rag}. However, these approaches treat all items as a single corpus, ignoring inherent object features for multi-source retrieval. To address this, we introduce MoK-RAG, the first multi-source RAG agent system.  

\paragraph{Embodied AI Environments Generation} 
Previous works rely on 3D artists for environment design~\citep{9157346, Gan2020ThreeDWorldAP, khanna2024habitat, kolve2017ai2, li2023behavior, puig2018virtualhome, xia2018gibson}, which limits scalability. Some methods construct scenes from 3D scans~\citep{ramakrishnanhabitat, savva2019habitat, szot2021habitat}, but these lack interactivity. Procedural frameworks like PROCTHOR~\citep{deitke2022} and Phone2Proc~\citep{deitke2023phone2proc} generate scalable environments. HOLODECK~\citep{holodeck} is a system that
generates 3D environments to match a user-supplied prompt. However, these approaches retrieve 3D objects from a single corpus without leveraging object relationships. MoK-RAG3D addresses this by utilizing multi-source retrieval to enhance contextual coherence.

\section{Methodology}
\begin{figure*}[t!]
    \centering
    \includegraphics[height=0.565\textwidth]{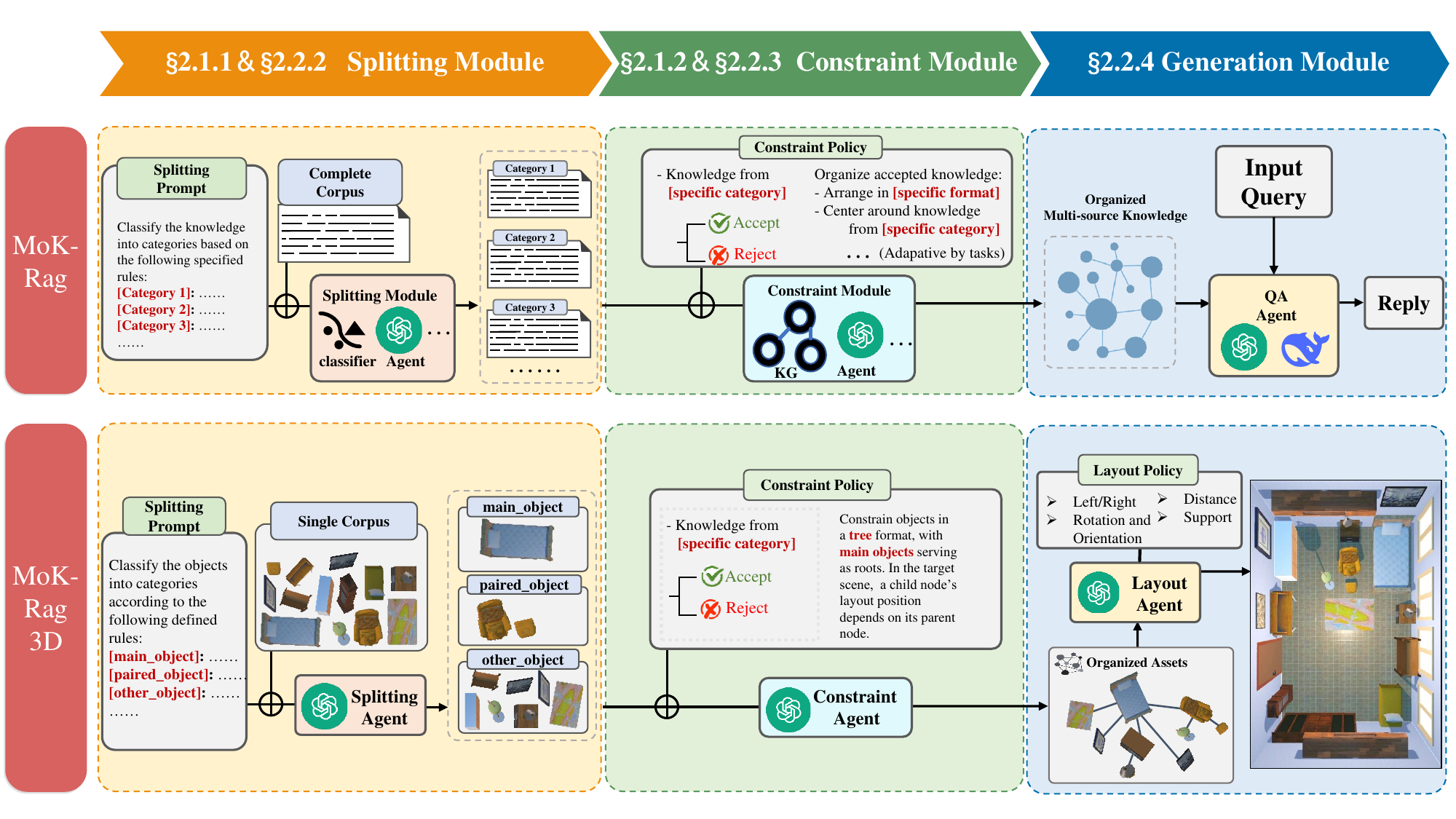}
          \caption{Overview of MoK-RAG and MoK-RAG3D. MoK-RAG consists of Splitting and Constraint modules for multi-source knowledge retrieval and Generation Module for response generation. MoK-RAG3D refines the Generation Module of the MoK-RAG framework into a dedicated Layout Module to facilitate scene generation.}
    \label{fig:framework}
\end{figure*}
\paragraph{Problem Formulation} In this paper, we study the problem of utilizing RAG for Embodied AI Environments. 
Existing RAG systems mainly focus on retrieving knowledge from a monolithic knowledge base. To advance this technique, we first explore how this design hinders the effectiveness of RAG systems.

In many cases, an ideal reply of RAG systems is structured and composed of multiple interdependent sections. However, traditional RAG systems, which rely on a single monolithic knowledge corpus, often fail to retrieve all necessary components, leading to responses that are incomplete or lacking key details. We refer to this problem as \textit{Reply Missing}.
For example, in a multimodal query answering task, an LLM may be asked to generate an image of a dragon along with a detailed caption. A traditional RAG framework, constrained to a single retrieval source, may only retrieve partial information—either the image or the caption—resulting in an incomplete response. 

Inspired by the decision-making processes of human cognition, retrieving information from diverse and specialized knowledge sources could be a promising direction.
Hence, our aim is to address this issue by leveraging multiple retrieval sources, ensuring each section of the reply is constructed from the most relevant knowledge base. 
For example, the new RAG design could retrieve the image from an image database while simultaneously retrieving the text description from a textual knowledge source, thereby retrieving knowledge from diverse knowledge paths and completing the structured response.

To this end, we propose MoK-RAG to achieve retrieving knowledge through diverse knowledge paths and propose MoK-RAG3D to adapt MoK-RAG to the problem of generation of 3D simulated environments.
As illustrated in Figure~\ref{fig:framework}, the MoK-RAG framework consists of three key components: \textbf{a Splitting Module}, which partitions a knowledge base into multiple knowledge paths,  \textbf{a Constraint Knowledge Module}, which organizes the retrieved knowledge, and a Generation Module to generate reply. MoK-RAG3D refines the Generation Module of the MoK-RAG framework into a dedicated \textbf{Layout Module} to facilitate scene generation. Next, we will illustrate the design of each module.
%In contrast, MoK-RAG addresses this issue by leveraging multiple retrieval sources, ensuring each section of the reply is constructed from the most relevant knowledge base. 

%A single retrieval source may lead to the omission of essential components in generated responses, a limitation we refer to as the ReplyMissing Problem. In this section, we first formally define the ReplyMissing Problem and then present the key modules of MoK-RAG.

%\paragraph{ReplyMissing Problem}  
%We found one problem in traditional RAG, which we call the \textit{ReplyMissing} problem. In many cases, a reply is structured and composed of multiple interdependent sections. However, traditional RAG systems, which rely on a single monolithic knowledge corpus, often fail to retrieve all necessary components, leading to responses that are incomplete or lacking key details. For example, in a multimodal query answering task, an LLM may be asked to generate an image of a dragon along with a detailed caption. A traditional RAG framework, constrained to a single retrieval source, may only retrieve partial information—either the image or the caption—resulting in an incomplete response. We refer to this problem as \textit{Reply Missing}. In contrast, MoK-RAG addresses this issue by leveraging multiple retrieval sources, ensuring each section of the reply is constructed from the most relevant knowledge base. Specifically, MoK-RAG retrieves the image from an image database while simultaneously retrieving the text description from a textual knowledge source, thereby completing the structured response. 

\subsection{MoK-RAG} 

%Shown in Figure~\ref{fig:framework}, the MoK-RAG framework is composed of three main components: the QA agent, the splitting agent, and the constraint agent. While the QA agent is responsible for generating responses by retrieving and synthesizing relevant knowledge, the splitting agent and constraint agent focus on preprocessing and organizing the knowledge base. 

%Mathematically, the structured retrieval process can be expressed as:  
%\begin{equation}
%    \text{ReplyMissing} = \sum_{j=1}^{m} M(q, k_j), \quad k_j \in D_j
%\end{equation}
%where each \( D_j \) represents a distinct knowledge section contributing to the final response.  
%By explicitly separating retrieval sources and enforcing structured knowledge integration, MoK-RAG effectively resolves the \textit{ReplyMissing} problem, ensuring comprehensive and coherent responses in complex generation tasks.
\paragraph{Splitting Module} 
The core insight of MoK-RAG lies in retrieving knowledge from diverse knowledge paths. Hence the key is to 
segment the knowledge base \( K \) into multiple specialized knowledge bases \( K_1, K_2, ..., K_m \), each base aligned with a specific domain or contextual theme. 

To achieve this goal, a dedicated splitting module is employed to partition the knowledge base. This module can be implemented using a classifier or an LLM-based agent, depending on the specific task requirements. The category set can be predefined or dynamically determined based on the characteristics of the task. Notably, when the number of categories is set to one, MoK-RAG degenerates into a conventional RAG system, making it a more generalized framework. 

Formally, given a knowledge base \( K = \{ k_1, k_2, ..., k_n \} \) containing \( n \) knowledge pieces and a category set \( C = \{ c_1, c_2, ..., c_m \} \), the objective of the splitting module is to assign each knowledge piece \( k_i \) to an appropriate category \( c_j \). This process can be represented as:
%\begin{equation}
%    f_{\text{split}}: k_i \to c_j, \quad \text{where } c_j = \arg\max_{c \in C} \text{AScore}(k_i, c)
%\end{equation}
\begin{equation}
\begin{aligned}
& f_{\text{split}}: k_i \to c_j, \quad \\ \text{where } c_j = &\arg\max_{c \in C} \text{AlignmentScore}(k_i, c)
\end{aligned}
\end{equation} 
where \( f_{\text{split}} \) denotes the splitting function, and \( \text{AlignmentScore}(k_i, c) \) is a relevance function that evaluates the alignment between \( k_i \) and category \( c \). 
\paragraph{Constraint Module}  After segmenting the knowledge base into multiple sections, it becomes essential to not only retrieve relevant knowledge but also to effectively organize the retrieved information into a structured form that enhances the generative capabilities of the LLM. While retrieval algorithms have been extensively studied in prior works~\cite{csakar2025maximizing}, we focus here on the organization of retrieved knowledge from multiple knowledge bases. This process can be categorized into two key aspects:

\textbf{Access Policy.} This policy determines whether a retrieved knowledge piece from a specific knowledge base should be accepted or rejected for inclusion in the final output. Formally, given a retrieved knowledge set \( K' = \{ k'_1, k'_2, ..., k'_l \} \), an access function \( f_{\text{access}} \) is defined as:
    \begin{equation}
        f_{\text{access}}(k'_i) = 
        \begin{cases} 
            1, & \text{if } \text{SelectionScore}(k'_i) \geq \tau \\
            0, & \text{otherwise}
        \end{cases}
    \end{equation}
    where \( \text{SelectionScore}(k'_i) \) represents the relevance score of \( k'_i \), and \( \tau \) is a predefined threshold controlling knowledge selection.

\textbf{Knowledge Organization Policy.} This policy defines the structural arrangement of the final knowledge representation. For instance, if the final output is a hierarchical knowledge tree, the organization algorithm must determine the placement of different nodes and their interrelations. Formally, given a hierarchical knowledge representation \( T = (N, E) \), where \( N \) denotes the set of nodes (knowledge units) and \( E \) represents the edges (relationships), the organization function \( f_{\text{org}} \) assigns retrieved knowledge pieces to appropriate nodes:
   \begin{equation}
\begin{aligned}
    f_{\text{org}}(k'_i) &\to n_j, \quad  \\ \text{where } n_j \in N,  \quad
    &\text{Relation}(n_j, n_k) \in E.
\end{aligned}
\end{equation} 

By enforcing these constraints, MoK-RAG can effectively refines the retrieved information, ensuring both the quality and structured coherence of the knowledge used for generation.

\subsection{MoK-RAG3D} 
%Shown in Figure 2, the MoK-RAG3D framework is composed of three main components: the QA agent, the splitting agent, and the constraint agent. While the QA agent is responsible for generating responses by retrieving and synthesizing relevant knowledge, the splitting agent and constraint agent focus on preprocessing and organizing the knowledge base.
Due to the complex requirements of 3D environment generation, the \textit{Reply Missing} Problem occurs more frequently in this domain. In this section, we first analyze its occurrence rate and then introduce the key modules of MoK-RAG3D.
\paragraph{The Occurance Rate of Problem \textit{Reply Missing}}
The task of 3D environment creation involves generating realistic virtual spaces based on textual descriptions. Within this task, we define two critical categories of objects: main objects and paired objects. Main objects are the central elements of an environment, without which the scene cannot be functionally or semantically complete. For instance, in a living room, a sofa serves as the main object—without it, the room loses its defining purpose. Similarly, in a bedroom, the bed is indispensable. On the other hand, paired objects refer to elements that frequently appear together within the same context, reinforcing semantic and functional coherence. Examples include a pot and a stove in a kitchen or a monitor and a keyboard in an office setting.  

We observed that the \textit{Reply Missing} problem frequently occurs in the 3D environment generation task. This problem manifests as missing essential objects, either main objects or paired objects, leading to incomplete or inconsistent environments. A missing main object results in a failure to establish the fundamental identity of the environment, while a missing paired object disrupts the expected co-occurrence patterns, reducing realism and usability.  

\begin{figure}[htbp]
    \centering \includegraphics[height=0.25\textwidth]{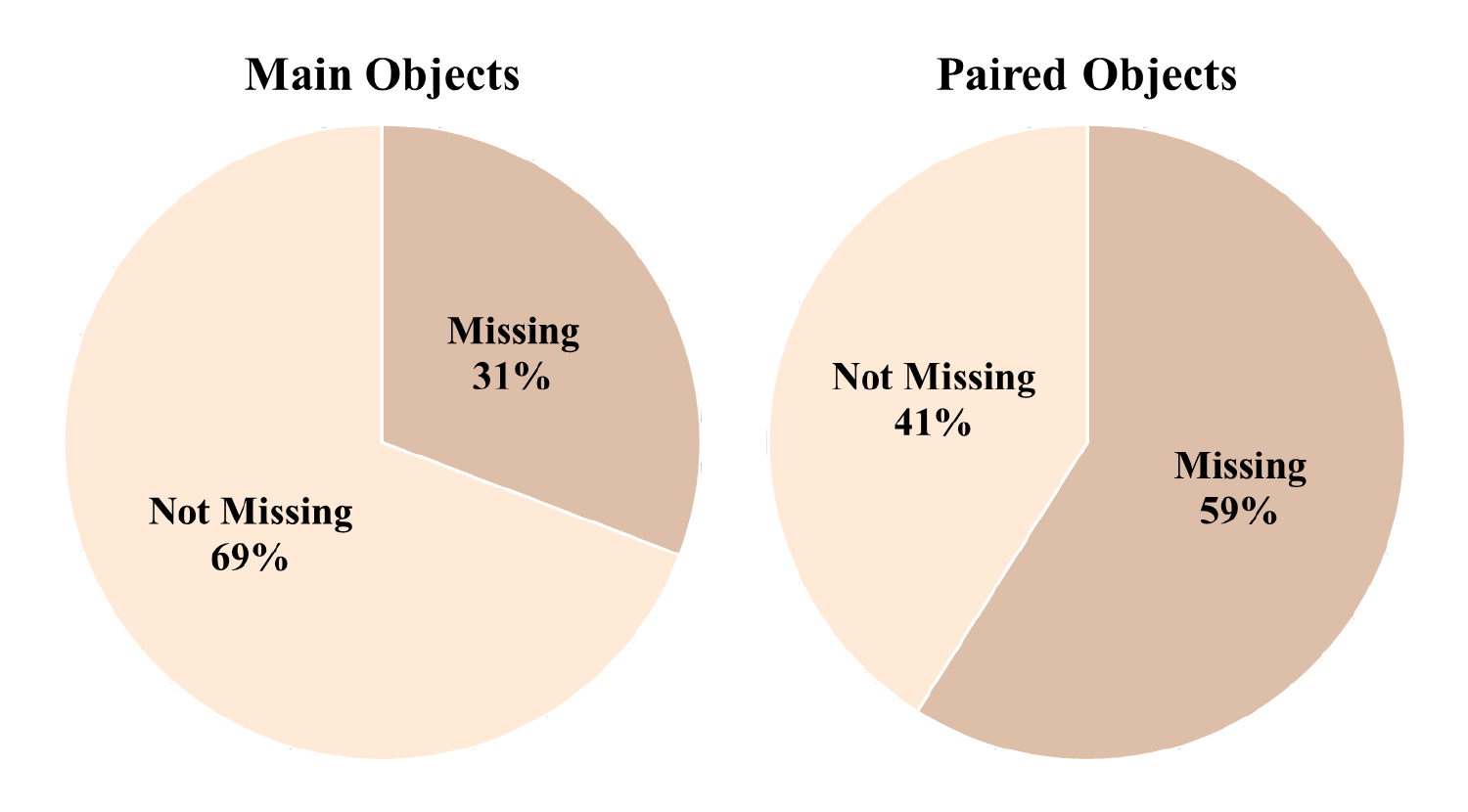}
          \caption{The Occurance Rate of Problem \textit{Reply Missing} from main objects and paired objetcs two aspects. }
    \label{fig:preExperiment}
\end{figure}

To quantify this issue, we conducted an empirical study where we generated 100 3D environment samples using a standard procedural generation approach. Each generated environment was then annotated by human experts to identify missing main objects and paired objects. The results, presented in Figure~\ref{fig:preExperiment}, indicate that 31\% of environments lack their main objects, while a significantly higher 59\% of paired objects are missing.  

These findings highlight the severity of the \textit{Reply Missing} problem in 3D environment creation. The high occurrence rate of missing elements underscores the urgent need for an improved retrieval and generation mechanism. Addressing this issue is essential for ensuring the completeness, realism, and functional integrity of automatically generated 3D environments.

\paragraph{Splitting Module of MoK-RAG3D} Following the structural design of MoK-RAG, the splitting module in MoK-RAG3D is implemented using an LLM-based agent. To address the \textit{Reply Missing} problem in 3D environment generation, the 3D object base is partitioned into three distinct sections: the \textit{main objects base}, the \textit{paired objects base}, and the \textit{other objects base}. 

Formally, given a 3D object knowledge base \( O = \{ o_1, o_2, ..., o_n \} \) consisting of \( n \) objects, and a category set \( C = \{ c_{\text{main}}, c_{\text{paired}}, c_{\text{other}} \} \), the splitting module assigns each object \( o_i \) to the most appropriate category \( c_j \). This can be formulated as:
\begin{equation}
\begin{aligned}
    & f_{\text{split}}: o_i \to c_j, \quad \\ \text{where } c_j = &\arg\max_{c \in C} \text{AlignmentScore}(o_i, c) 
\end{aligned}
\end{equation}
where \( f_{\text{split}} \) represents the splitting function, and \( \text{AlignmentScore}(o_i, c) \) is a scoring function that quantifies the relevance of object \( o_i \) to category \( c \). 
%By structuring the 3D object base in this manner, MoK-RAG3D ensures a more accurate and context-aware retrieval process, effectively mitigating missing object issues in generated 3D environments.

\paragraph{Constraint Module of MoK-RAG3D} As illustrated in Figure~\ref{fig:framework}, the splitting agent partitions the knowledge base into three sections: the main objects base, the paired objects base, and the other objects base, denoted as \( C = \{ c_{\text{main}}, c_{\text{paired}}, c_{\text{other}} \} \).  
The constraint module in MoK-RAG3D follows the fundamental structure of MoK-RAG and consists of two crucial aspects:  

\textbf{Access Policy:} After retrieving the most relevant objects from different knowledge bases using LLM Agent, an access policy is applied to filter and refine the retrieved objects.  Formally, given a set of retrieved objects \( O = \{ o_1, o_2, ..., o_k \} \) from multiple knowledge bases, the access function \( f_{\text{access}} \) operates as follows:  
    \begin{equation}  
        O_{\text{filtered}} = f_{\text{access}}(O),  
    \end{equation}  
    where \( O_{\text{filtered}} \) represents the final selection of objects after filtering redundant or irrelevant elements.  

\textbf{Knowledge Organization Policy:} The retrieved knowledge is structured into a hierarchical tree using a multi-round querying strategy of LLM. Specifically, given the retrieved main objects set \( M = \{ m_1, m_2, ..., m_p \} \), the paired objects set \( P = \{ p_1, p_2, ..., p_q \} \), and the other objects set \( O = \{ o_1, o_2, ..., o_r \} \), the hierarchical organization follows these steps:  
\begin{itemize}  
        \item Root Node Decision: The main objects from \( M \) are selected as the root nodes of the hierarchical tree.  
        \item Node Hierarchy Determination: The LLM is iteratively queried to determine the child nodes for each parent node. This process continues recursively until all objects from \( P \) and \( O \) are assigned appropriate positions within the tree. Ultimately, this results in the construction of multiple hierarchical trees, where the number of trees corresponds to the number of main objects in \( M \).  
\end{itemize}  
This hierarchical organization ensures that the retrieved knowledge is systematically structured, allowing for a coherent and contextually rich 3D environment generation.  

\paragraph{Layout Module of MoK-RAG3D}  
After constructing the 3D layout tree, it is crucial to establish the spatial relationships between different objects to ensure a coherent scene structure. We define four key relationship categories as follows:  

\begin{itemize}  
    \item Left/Right: This relationship specifies the relative horizontal positioning of objects, determining whether an object is placed to the left or right of another object.  
    \item Rotation and Orientation: This aspect defines the angular alignment of objects, ensuring that they are correctly rotated to fit the intended scene context.  
    \item Distance: This relationship governs the spatial separation between objects, maintaining a realistic distribution of objects within the environment.  
    \item Support: This category captures structural dependencies, ensuring that objects requiring support (e.g., a book on a table) are correctly positioned with respect to their supporting surfaces.  
\end{itemize}  

To determine these relationships, we iteratively query the LLM for each of the four defined relationships along every edge in the layout tree. During the layout process, the position of the root node (i.e., the main object) is first determined. Subsequently, child nodes are placed iteratively by considering their respective relationships with their parent nodes, ensuring a structurally consistent and semantically meaningful 3D environment.  
%Edge Relationship Definition. Establish the edges between nodes by defining the specific relationships (e.g., ) between parent and child nodes. Assign attributes or weights to edges if necessary to encode additional hierarchical information.
%\input{4-experimental}
%\section{Results and Discussion}
%\section{Human Evaluation}
%We conduct comprehensive human evaluations to assess the quality of MoK-RAG3D scenes, with a total of 120 participants in three user studies: (1) a comparative analysis on residential scenes with HOLODECK as the baseline; (2) an examination of MoK-RAG3D’s ability in generating diverse scenes. Through these user studies, we demonstrate that MoK-RAG3D can create residential scenes of better quality than previous work while being able to extend to a wider diversity of scene types.
\section{Experiments}
\subsection{Experimental Setup}
\paragraph{Datasets.}
We utilize Objaverse 1.0~\citep{objaverse}, a large-scale dataset containing over 800,000 3D models, as the source for object selection in 3D environment construction. Following~\citet{holodeck}, we evaluate our LLM agents across two categories: residential and diverse scenes. The residential scenes include bathroom, bedroom, kitchen, and living room. For diverse scenes, we use the MIT Scenes Dataset~\citep{mitscene}, which provides the largest available collection of indoor scene categories across various domains.  

\paragraph{Metrics.}
Following~\citet{holodeck}, we conduct large-scale human evaluations to assess the quality of generated 3D environments. Annotators rate the scenes on a scale of 1 to 5 based on asset selection, layout coherence, and overall alignment with the intended scene type. Additionally, inspired by~\citet{gpt4oeval}, we introduce an automated evaluation method for 3D environment generation. Specifically, we leverage different LLMs as evaluators to provide an objective assessment of the generated scenes.  

\paragraph{Models.} 
MoK-RAG3D comprises three core components: the splitting agent, the constraint agent, and the QA agent, all implemented using GPT-4-1106-preview~\cite{achiam2023gpt}. In our current implementation, MoK-RAG3D can generate a single room in approximately three minutes, including the time required for API calls and layout optimization. All experiments are conducted on a MacBook equipped with an M1 chip.  

\subsection{Human Evaluation}  
To assess the quality of MoK-RAG3D-generated scenes, we conduct comprehensive human evaluations involving 120 participants across two user studies:  
(1) a comparative analysis of residential scene generation; 
(2) an evaluation of MoK-RAG3D's capability in generating diverse scenes.  %These studies demonstrate that MoK-RAG3D not only produces higher-quality residential scenes compared to prior work but also generalizes effectively to a broader range of scene types. 
\paragraph{Residential Scenes Evaluation.}
We conducted a human evaluation with 120 generated scenes, evenly distributed across four residential scene types (30 scenes per type) for both MoK-RAG3D and the HOLODECK baseline. Both systems utilized the same Objaverse asset set to ensure a fair comparison.  

For MoK-RAG3D, we provided the scene type (e.g., “bedroom”) as the input prompt for scene generation. Scenes of the same type from both systems were paired, resulting in 120 matched scene pairs. Each pair was presented to annotators as two shuffled top-down view images, ensuring that the generating system remained anonymous.  

Annotators were asked to evaluate each scene based on three key criteria:  
(1) Asset Selection: Which system selects 3D assets that are more accurate and faithful to the scene type?  
(2) Layout Coherence: Which system arranges 3D assets in a more realistic and logically consistent manner (considering position and orientation)?  
(3) Overall Preference: Given the scene type, which scene is preferred overall?  

\begin{figure}[htbp]
    \centering \includegraphics[height=0.17\textwidth]{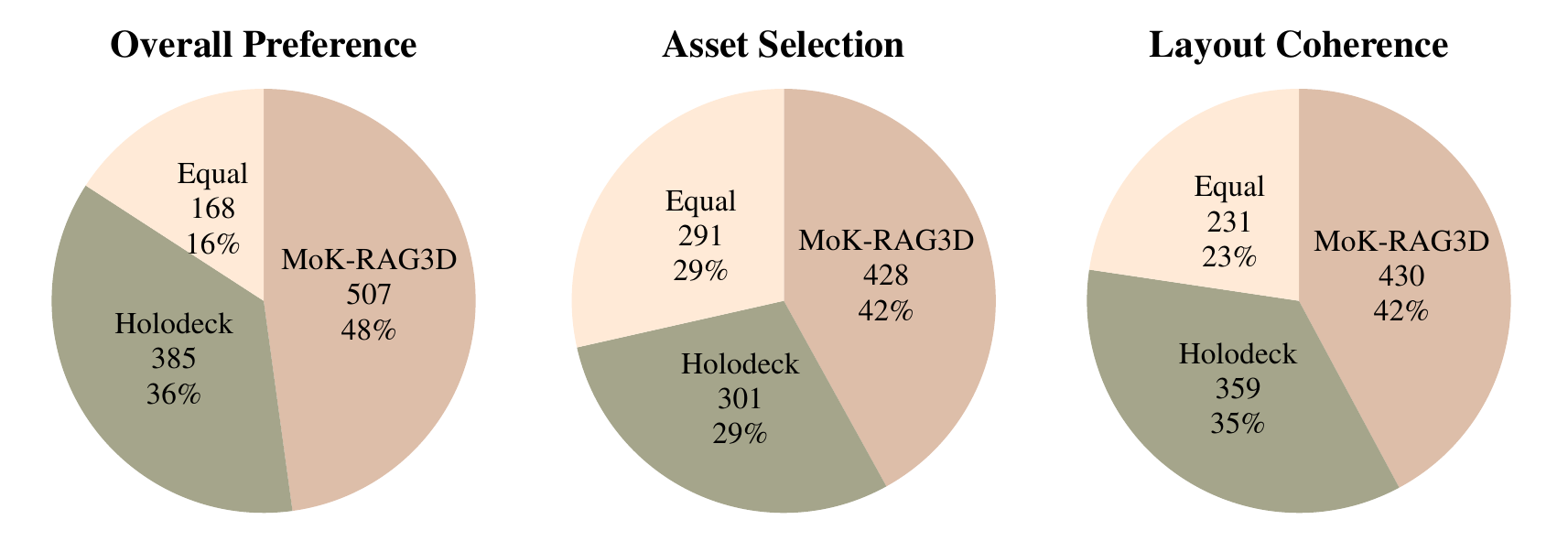}
          \caption{Comparative human evaluation of MoK-RAG3D and HOLODECK across three criteria. The pie charts show the distribution of annotator preferences, showing both the percentage and the actual number of annotations favoring each system.}
    \label{fig:exp1}
\end{figure}

%\textbf{Human Preferences for MoK-RAG3D}  
Figure~\ref{fig:exp1} shows a clear preference for MoK-RAG3D in human evaluations compared to HOLODECK. Annotators favored MoK-RAG3D in Asset Selection (42\%), Layout Coherence (42\%), and demonstrated a significant preference in Overall Preference (48\%). These results indicate that MoK-RAG3D produces more realistic and semantically appropriate 3D environments.  

\begin{figure*}[htbp]
    \centering \includegraphics[height=0.58\textwidth]{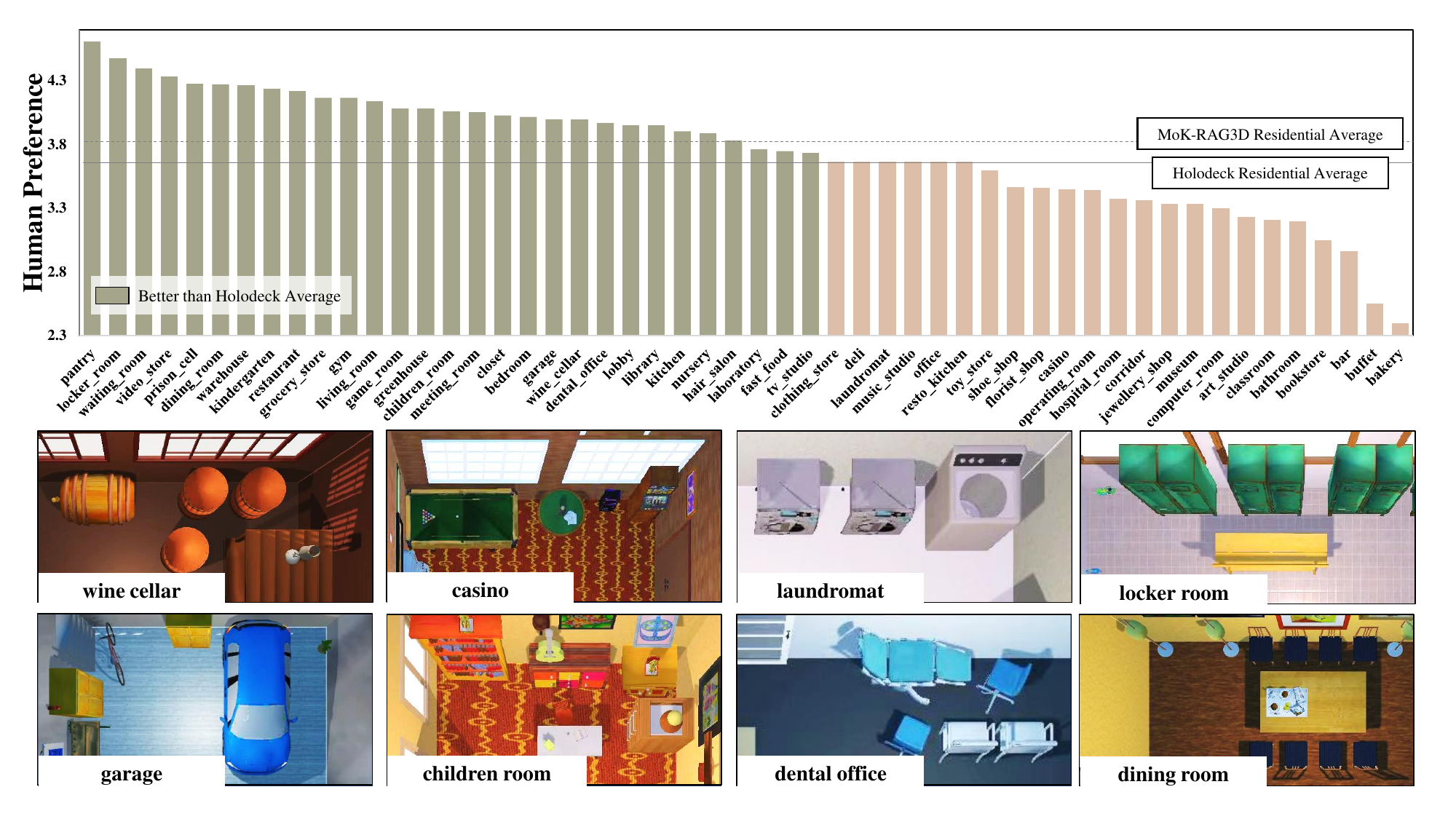}
          \caption{Human evaluation on 52 scene types from MIT Scenes Dataset~\citep{mitscene} with qualitative examples. The two horizontal lines represent the
average score of MoK-RAG3D and HOLODECK on four types of residential scenes (bedroom, living room, bathroom and kitchen.) }
    \label{fig:exp2}
\end{figure*}

\paragraph{Scenes Diversity Analysis.}
To assess MoK-RAG3D’s performance beyond residential scenes, we conducted a human evaluation on 52 scene types from the MIT Scenes Dataset, covering five categories: Stores (deli, bakery), Home (bedroom, dining room), Public Spaces (museum, locker room), Leisure (gym, casino) and Working Space (office, meeting room). We prompt MoK-RAG3D to produce five outputs for each type using only the scene name as the input, accumulating 260 examples across the 52 scene types. Annotators are presented with a top-down view image and a 360-degree video for each scene and asked to rate them from 1 to 5 (with higher scores indicating better quality), considering
asset selection, layout coherence, and overall match with the scene type. 
%\textbf{MoK-RAG3D can generate satisfactory outputs for most scene types} 
Figure ~\ref{fig:exp2} demonstrates the human preference scores for diverse scenes with qualitative examples. Compared to SpiltRagFor3D’s performance in residential scenes, SpiltRagFor3D achieves higher human preference scores over half of (29 out of 52) the diverse scenes.

\subsection{Automatic Evaluation}
\paragraph{Automatic Evaluation on generated 3D scenes.}
%\paragraph{Setup}  
%The automatic evaluation of 3D environment generation quality remains an underexplored area. 
To assess the quality of the generated environments, we employ two evaluation models: (1) GPT-4o, a closed-source model that shares the same origin as the LLM agents used in our system; and (2) LLaVA, an open-source model known for its strong multi-modal comprehension capabilities.  

To facilitate evaluation, we transform each generated 3D environment into a sequence of four images by rotating the scene every 90 degrees. These images are then fed into the evaluation models. The evaluation models are asked to  rate them from 1 to 5 (with higher scores indicating better quality), considering
asset selection, layout coherence, and overall match
with the scene type.

\begin{figure*}[htbp]
    \centering \includegraphics[height=0.32\textwidth]{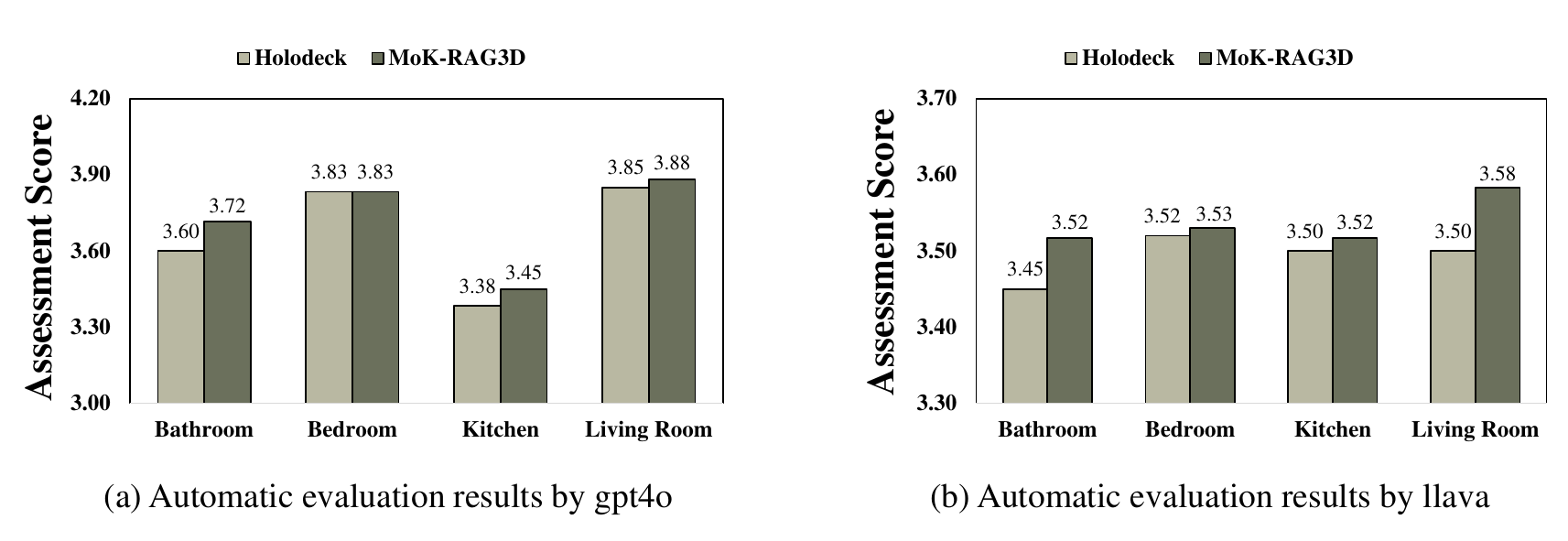}
          \caption{Automatic evaluation results comparing Holodeck and MoK-RAG3D across residential scenarios. The results from both GPT-4o and Llava consistently indicate MoK-RAG3D outperforms Holodeck in all evaluated environments. }
    \label{fig:exp3}
\end{figure*}

%\paragraph{Results}  
% presents a comparative analysis of MoK-RAG3D and Holodeck across different scene categories.
Shown in Figure~\ref{fig:exp3}, the evaluation results from both models in all residential scenarios consistently indicate that MoK-RAG3D outperforms Holodeck in most environments. These results highlight MoK-RAG3D’s overall advantage in generating high-quality 3D environments.

\begin{figure}[htbp]
    \centering \includegraphics[height=0.275\textwidth]{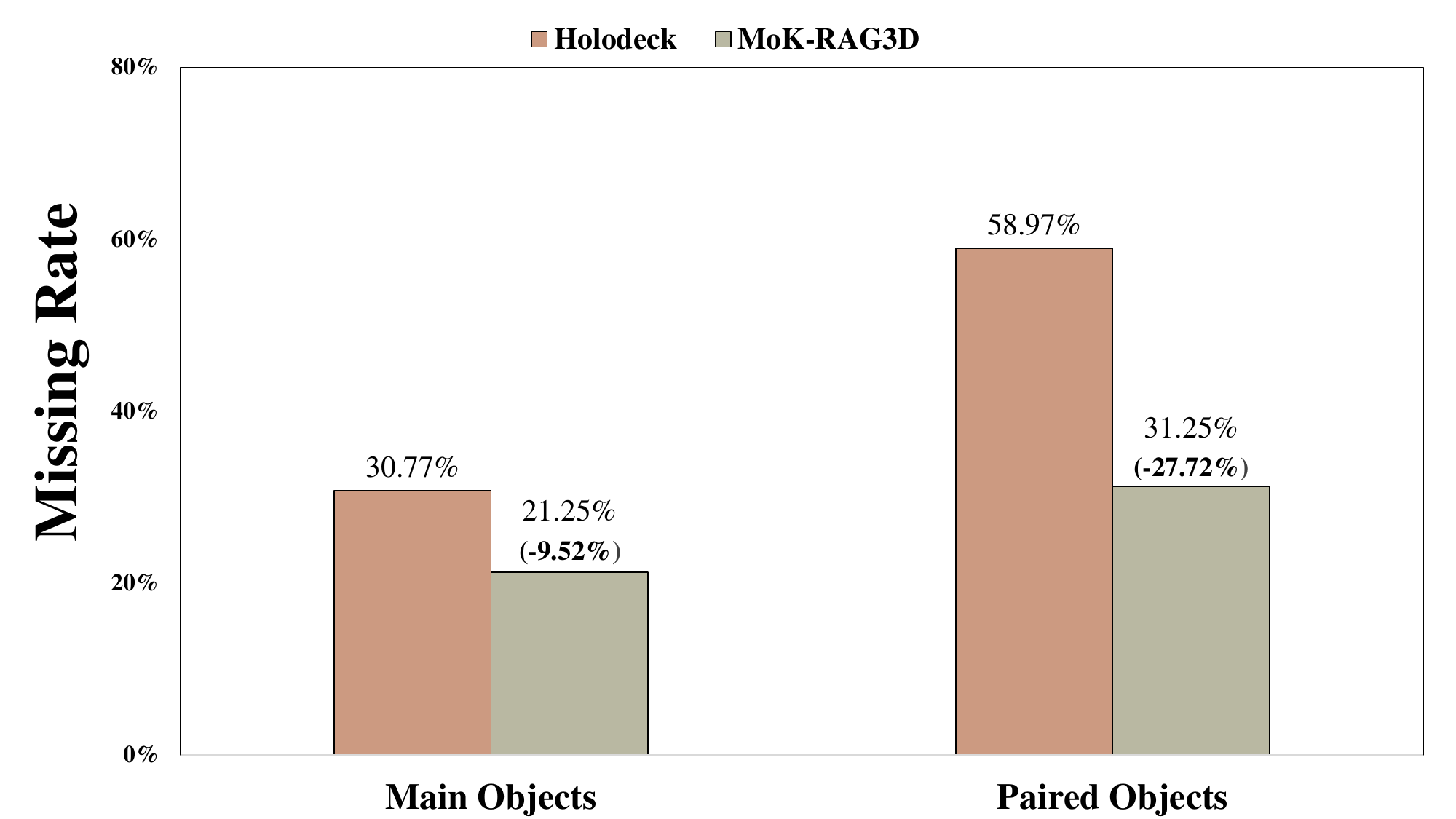}
          \caption{Comparison of missing rates between Holodeck and MoK-RAG3D for main objects and paired objects. MoK-RAG3D significantly reduces the missing rate. }
    \label{fig:exp6}
\end{figure}

\paragraph{Effectiveness evaluation(\textit{Reply Missing}).}
%\paragraph{Reduction of the occurance of \textit{Reply Missing} problem by MoK-RAG3D} 
%Due to MoK-RAG3D's ability to utilize multiple sources, it can effectively control the content associated with specific features, which indicates it has a significant advantage in addressing the issue of missing objects. 
MoK-RAG3D's multi-source retrieval enables precise control over feature-specific content, offering a distinct advantage in mitigating \textit{Reply Missing}. Figure~\ref{fig:exp6} illustrates that incorporating the MoK-RAG3D method leads to a substantial reduction in the missing rate, with a decrease of 9.52\% for main objects and 27.22\% for paired objects.

\begin{figure}[htbp]
    \centering \includegraphics[height=0.27\textwidth]{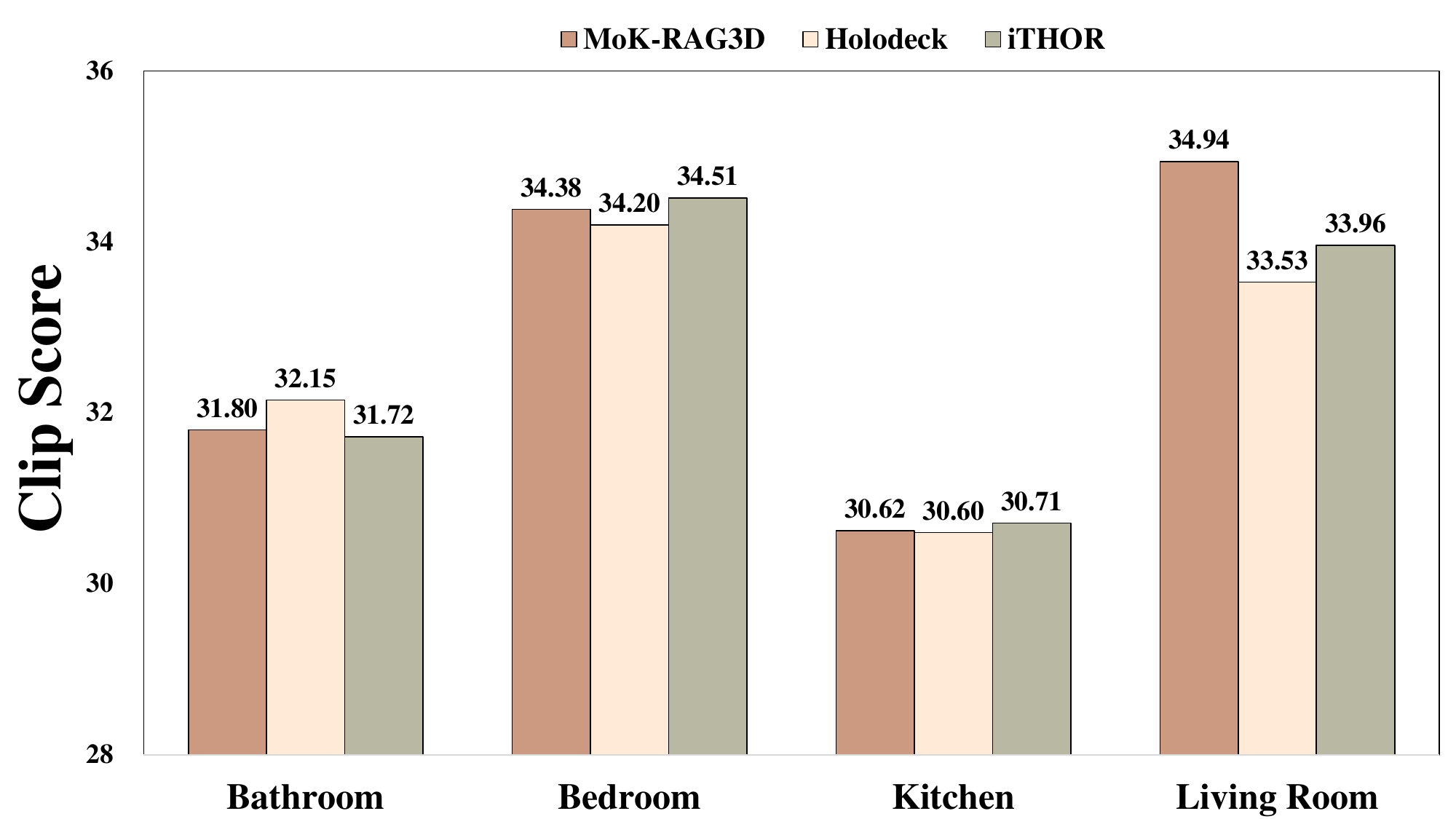}
          \caption{CLIP Score comparison over four residential scene types. * denotes iTHOR scenes are designed by human experts. }
    \label{fig:exp4}
\end{figure}

\paragraph{Effectiveness evaluation(scene quality).}
We use CLIP Score to assess the visual coherence between the top-down view of a generated scene and its designated scene type, following the prompt template: “a top-down view of [scene type].” Additionally, human-designed scenes from iTHOR serve as an upper bound for reference. As shown in Figure~\ref{fig:exp4}, MoK-RAG3D outperforms HOLODECK in most scenarios and closely approaches iTHOR, demonstrating its ability to generate scenes comparable to human-beings.
% with structural and visual fidelity
%Notably, the CLIP Score results are consistent with our human evaluation findings.
\begin{figure}[t]
    \centering
    \includegraphics[width=0.46\textwidth]{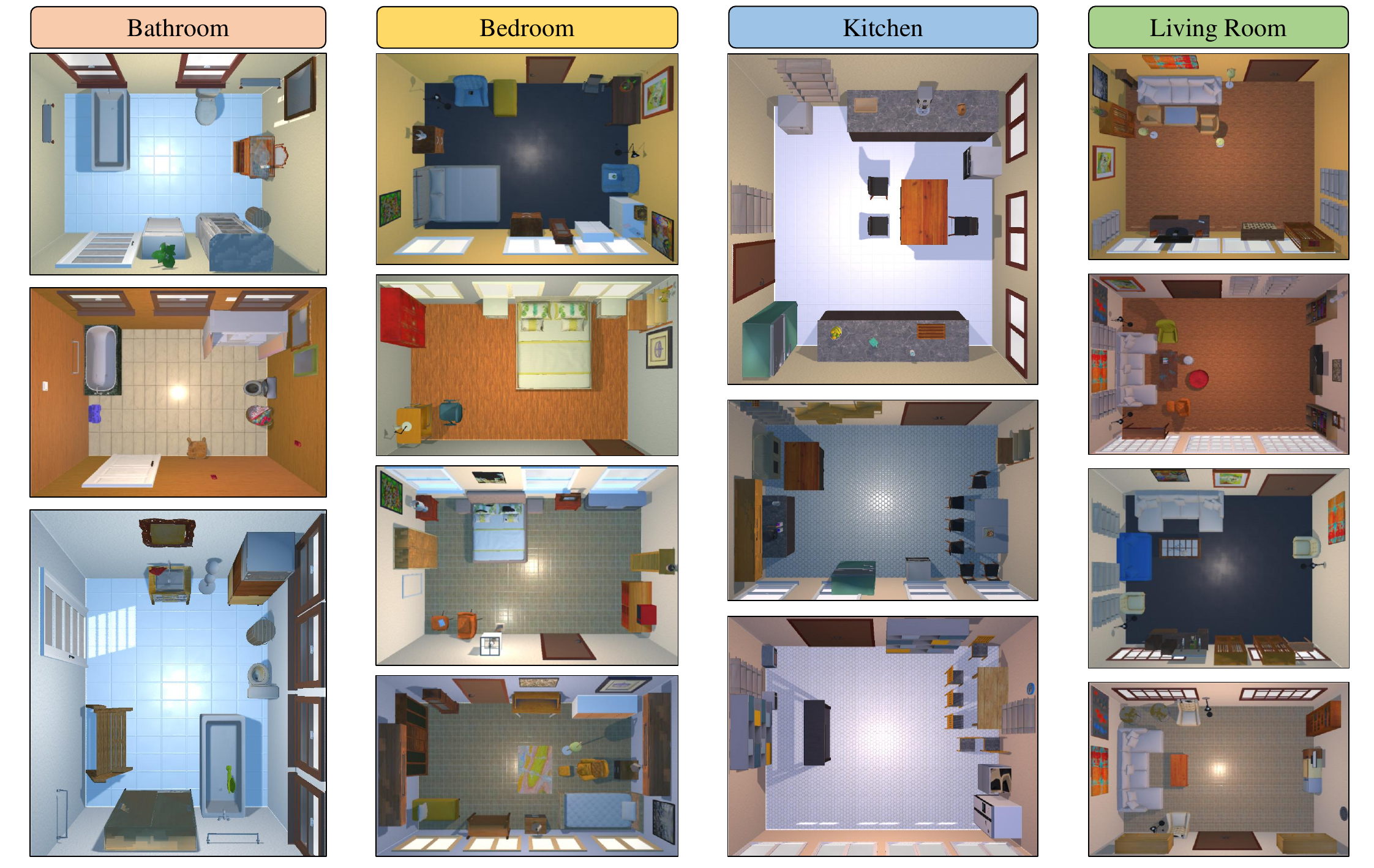}
    \caption{one example of residential scene results.}
    \label{fig:residential}
\end{figure}
\subsection{Visual Results}
%Figure \ref{fig:residential} presents a residential scene example demonstrating intuitive main object-centric zoning, highlighting the effectiveness of MoK-RAG3D. Additional visual results are available in the Appendix.
Figure \ref{fig:residential} is one residential scenes example exhibiting intuitive main objects-centric zoning, which can prove the effectiveness of Mok-RAG3D.
More visual results can be seen in Appendix.
\section{Conclusion}

%In this paper, we present an multi-source rag system, MoK-RAG and proposed its one application system Mok-RAG3D. It is the first RAG framework to enable multi-path knowledge retrieval by functionally partitioning LLM knowledge bases. Both automatic and human assessments confirming its effectiveness in enhancing Embodied AI agents' ability to generate diverse scenes.
%We introduce MoK-RAG, a multi-source RAG system, and its application, MoK-RAG3D. As the first RAG framework enabling multi-path knowledge retrieval through functional partitioning, it demonstrates effectiveness in enhancing Embodied AI agents' scene generation, as validated by both automatic and human evaluations.
In this paper, we introduce MoK-RAG, the first RAG framework enabling multi-path knowledge retrieval through functional partitioning of LLM knowledge bases, facilitating concurrent multi-source information retrieval. We further extend this framework to 3D environment generation with MoK-RAG3D, which improves scene realism and diversity. Additionally, MoK-RAG3D pioneers automated evaluation for 3D scene generation, with both automatic and human assessments validating its effectiveness in enhancing Embodied AI agents' ability to generate diverse scenes.

\newpage
\section{Limitations}
In this paper, MoK-RAG and MoK-RAG3D demonstrates excellent performance in enhancing Embodied AI agents’ ability to generate diverse scenes.
However, due to the lack of domain-specific hardware resource, it struggles with testing real robot in the generated scenes. This highlights the need for further enhancement to evaluation of the generated 3D scenes.

\bibliography{anthology,custom}
%\bibliography{anthology}
\bibliographystyle{acl_natbib}

%\newpage
\appendix
\section{APPENDIX}
%Use \verb|\appendix| before any appendix section to switch the section numbering over to letters. See Appendix~\ref{sec:appendix} for an example.

%\subsection{QUALITATIVE RESULTS OF RESIDENTIAL SCENES}

%Our method generates 3D scenes that balance aesthetic unity with context-aware object placement, ensuring layouts align with real-world spatial logic across diverse domains. As shown in \ref{fig:residentialScenes}, the generated residential scenes demonstrate intuitive activity-centric zoning and context-aware furniture arrangements.

\begin{figure*}[ht]
    \centering
    \includegraphics[width=\textwidth]{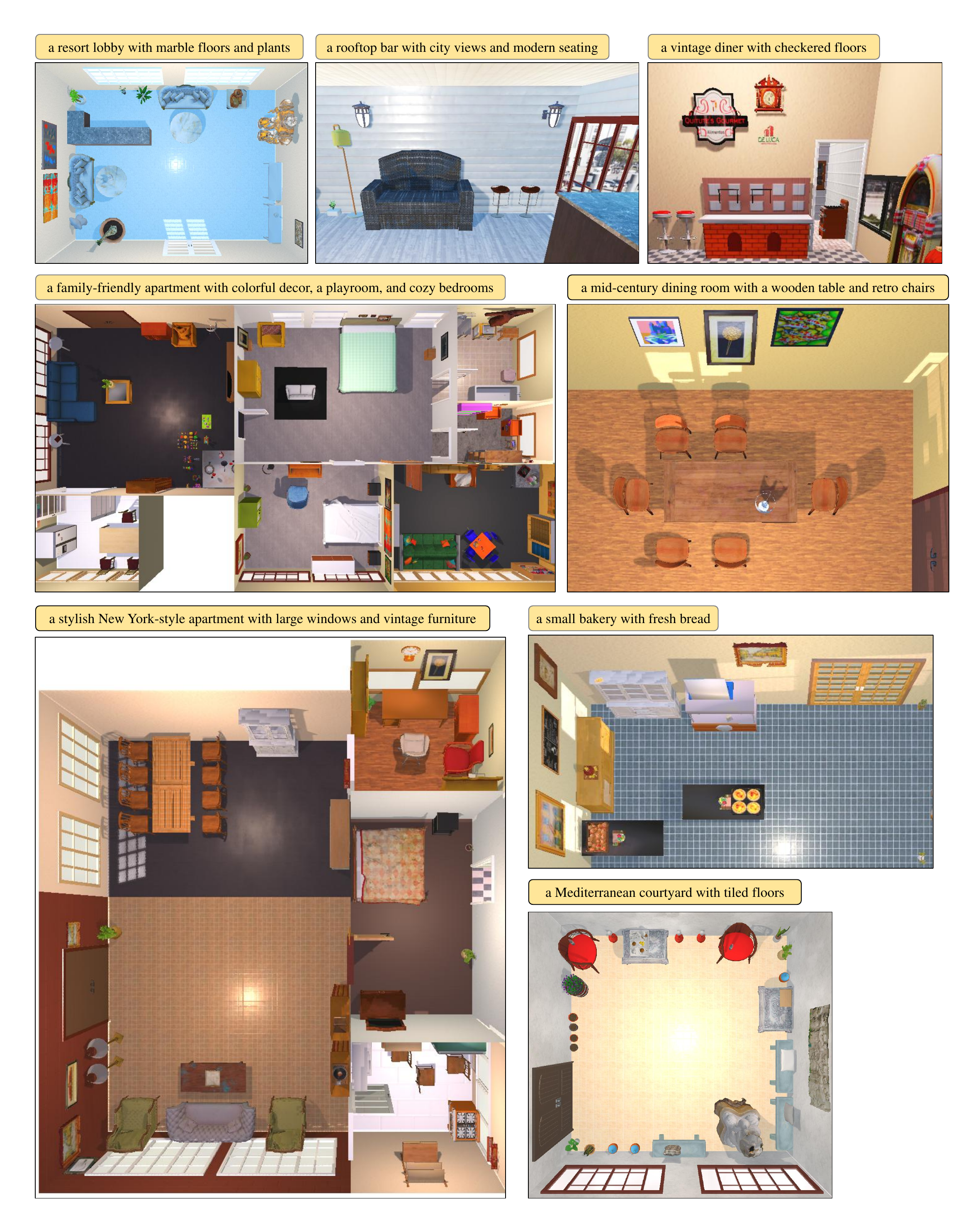}
    \caption{Some qualitative example of query-based scene results.}
    \label{fig:query-based}
\end{figure*}

\begin{figure*}[htbp]
    \centering
    \includegraphics[width=\textwidth]{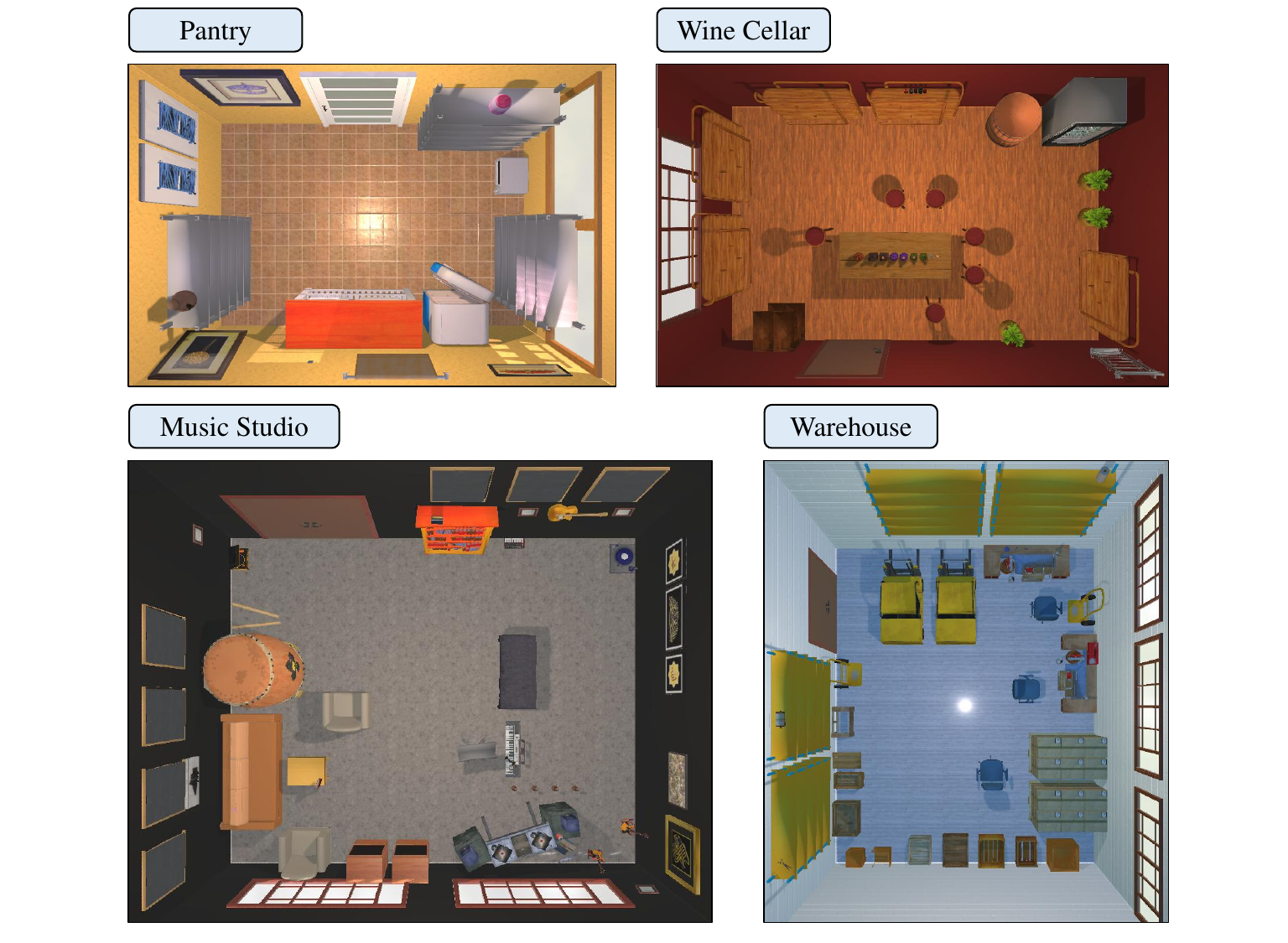}
    \caption{Some qualitative examples of results of scenes from MIT Indoor Scenes dataset.}
    \label{fig:residentialScenes}
\end{figure*}

%\subsection{MORE QUALITATIVE RESULTS }

%As shown in \ref{fig:MITscenes}, \ref{fig:query-based}, these scenes demonstrate our method's capacity to handle spatial requirements of diverse scenes. The generated scenes respect conventions without sacrificing natural object distribution, avoiding overly rigid layouts. Query-based scenes integrate semantic requirements with spatial rationalization.

\begin{figure*}[htbp]
    \centering
    \includegraphics[width=\textwidth]{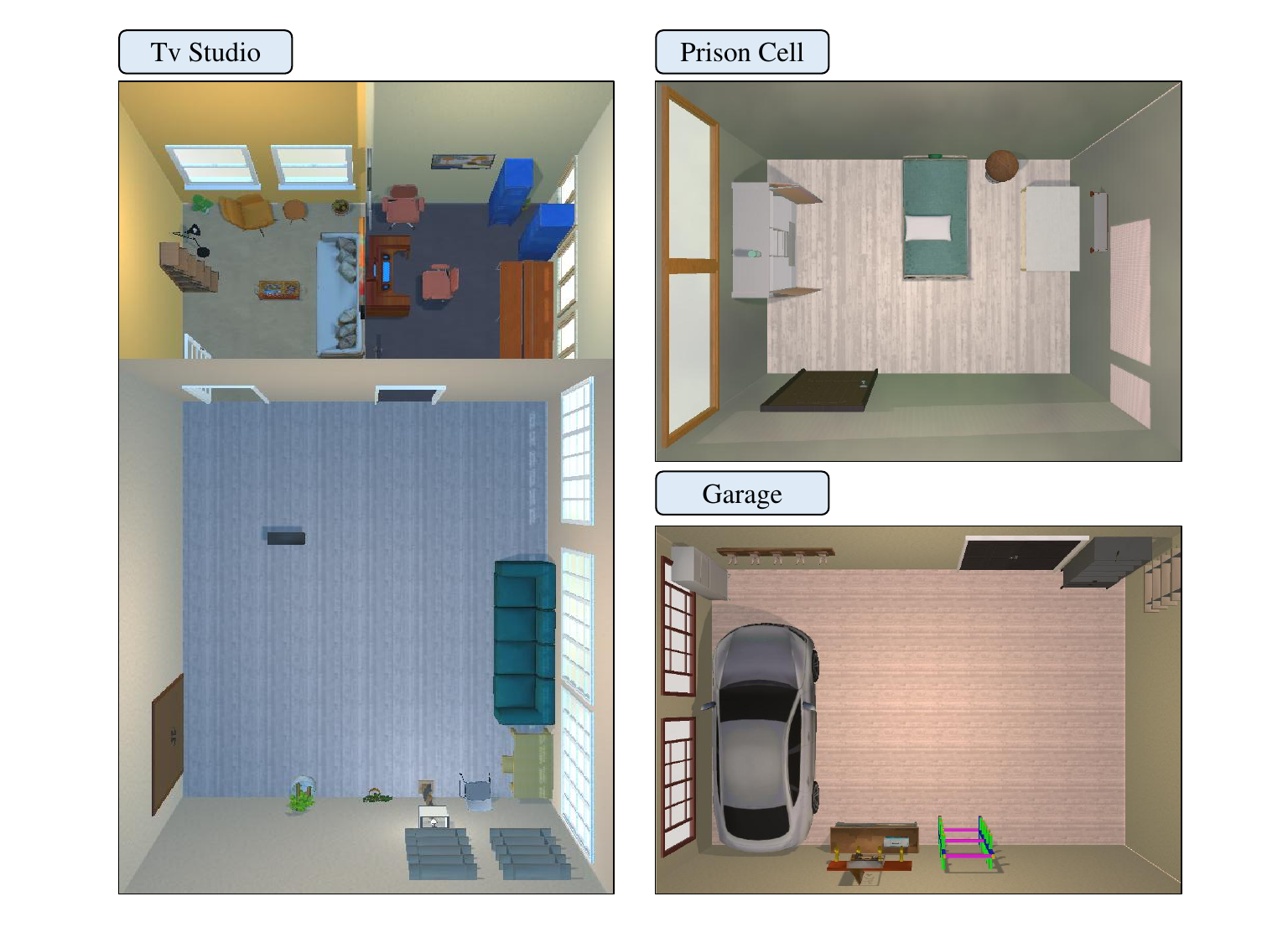}
    \caption{Some qualitative examples of results of scenes from MIT Indoor Scenes dataset.}
    \label{fig:MITscenes}
\end{figure*}

\end{document}